\documentclass[letterpaper]{article} 
\usepackage{aaai2026}  
\usepackage{times}  
\usepackage{helvet}  
\usepackage{courier}  
\usepackage[hyphens]{url}  
\usepackage{graphicx} 
\usepackage{url}
\usepackage{natbib}  
\usepackage{caption} 
\usepackage{algorithm}
\usepackage{algorithmic}

\usepackage{newfloat}
\usepackage{listings}
\usepackage{algorithm}
\usepackage{algorithmic}
\usepackage{amsmath}
\usepackage{amsfonts}
\usepackage{graphicx}
\usepackage{tabularx}
\usepackage{multirow}
\usepackage{kotex}
\usepackage{xcolor}
\usepackage{newfloat}
\usepackage{listings}
\usepackage{makecell}
\usepackage{booktabs}
\DeclareCaptionStyle{ruled}{labelfont=normalfont,labelsep=colon,strut=off} 
\floatstyle{ruled}
\newfloat{listing}{tb}{lst}{}
\floatname{listing}{Listing}
\title{DipGuava: Disentangling Personalized Gaussian Features for 3D Head Avatars from Monocular Video}

\author {
    Jeonghaeng Lee\textsuperscript{\rm 1},
    Seok Keun Choi\textsuperscript{\rm 1},
    Zhixuan Li\textsuperscript{\rm 2},
    Weisi Lin\textsuperscript{\rm 2},
    Sanghoon Lee\textsuperscript{\rm 1}
}
\affiliations {
    \textsuperscript{\rm 1}Yonsei University, Korea\\
    \textsuperscript{\rm 2}Nanyang Technological University, Singapore\\
    leedoright@yonsei.ac.kr, csg@yonsei.ac.kr, zhixuan.li@ntu.edu.sg, wslin@ntu.edu.sg, slee@yonsei.ac.kr
}

\usepackage{bibentry}

\begin{document}

\maketitle

\begin{abstract}
While recent 3D head avatar creation methods attempt to animate facial dynamics, they often fail to capture personalized details, limiting realism and expressiveness.
To fill this gap, we present \textbf{DipGuava} (Disentangled and Personalized Gaussian UV Avatar), a novel 3D Gaussian head avatar creation method that successfully generates avatars with personalized attributes from monocular video.
DipGuava is the first method to explicitly disentangle facial appearance into two complementary components, trained in a structured two-stage pipeline that significantly reduces learning ambiguity and enhances reconstruction fidelity.
In the first stage, we learn a stable geometry-driven base appearance that captures global facial structure and coarse expression-dependent variations.
In the second stage, the personalized residual details not captured in the first stage are predicted, including high-frequency components and nonlinearly varying features such as wrinkles and subtle skin deformations.
These components are fused via dynamic appearance fusion that integrates residual details after deformation, ensuring spatial and semantic alignment.
This disentangled design enables DipGuava to generate photorealistic, identity-preserving avatars, consistently outperforming prior methods in both visual quality and quantitative performance, as demonstrated in extensive experiments.
\end{abstract}


\section{1. Introduction}
\label{sec:intro}

\begin{figure}[t]
  \centering
  \includegraphics[width=0.47\textwidth]{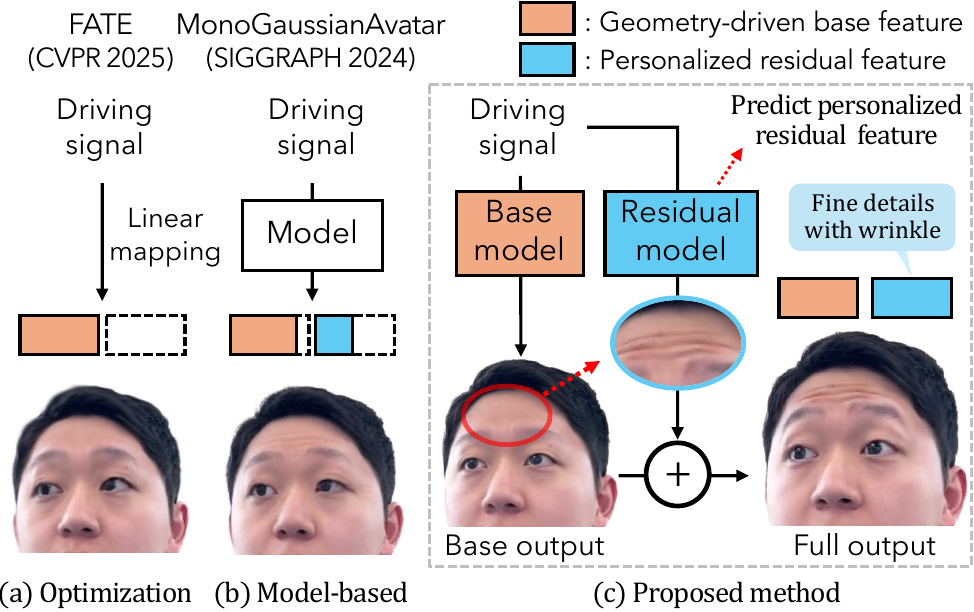}
\caption{Conceptual comparison with prior approaches. (a) Optimization-based methods fail to capture residual details. (b) Entangled models suffer from learning ambiguity. (c) Our disentangled design separately models base and residual features for faithful reconstruction.}
  \label{fig:main}
\end{figure}

Photorealistic 3D head avatars enable diverse applications such as VR, gaming, and telepresence, demanding efficient methods to create high-quality personalized avatars from monocular videos.
Recent advancements, particularly Neural Radiance Fields (NeRF)~\cite{mildenhall2021nerf} and 3D Gaussian splatting (3DGS)~\cite{kerbl20233d}, have enabled highly photorealistic and animatable head avatar creation. 
A common strategy is to use 3D Morphable Models (3DMMs)~\cite{paysan20093d, wang2022faceverse, gerig2018morphable, li2017learning} as a stable geometric prior and as a shared latent space for expression and pose, making them suitable for animating diverse subjects.
However, despite their progress, these approaches still struggle to capture the high-frequency, personalized attributes that define an individual’s unique appearance, often due to their reliance on coarse geometric priors or entangled representations. 
As a result, they fall short in reproducing the subject-specific details essential for truly personalized avatars.

We identify the root cause of these limitations by rethinking the shared problem formulation that underlies these methods.
As depicted in Fig.~\ref{fig:main}, we conceptualize facial features into two categories: \emph{(i)} geometry-driven base features, which are well-represented by a 3DMM that captures the average facial variations across many individuals (\emph{e.g.}, overall skin tone and rigid structural details); and \emph{(ii)} residual features that capture high-frequency, person-specific details such as wrinkles and subtle skin deformations not represented within the expressive range of 3DMMs.

Analyzing prior works through this disentangled perspective reveals the source of their shortcomings. 
Despite their architectural differences, most 3D head creation methods share the common, challenging formulation of mapping a low-dimensional set of 3DMM parameters to the complex, high-dimensional space of facial attributes.
Solving this problem with an optimization-based approach (Fig.\ref{fig:main}a, \cite{zhang2025fate}) can effectively capture stable base features via linear mappings from 3DMM parameters~\cite{zhao2024psavatar, qian2024gaussianavatars, shao2024splattingavatar, zhang2025fate}. 
However, their inherent linearity hinders modeling of non-linear residual details, even with subject-specific optimization.
In contrast, more recent feature prediction models (Fig.~\ref{fig:main}b, \cite{chen2024monogaussianavatar}) attempt to address these limitations using a model-based approach that learns expression-dependent feature deformation from canonical space~\cite{zielonka2023instant, kirschstein2023nersemble, xiang2024flashavatar, chen2024monogaussianavatar}. 
However, this holistic approach forces the network to learn both base and residual features simultaneously, a complex task that results in an entangled representation where the crucial, personalized details are suppressed.

To resolve these limitations, we propose the \textbf{DipGuava} (Disentangled and Personalized Gaussian UV Avatar), a novel framework that rethinks the problem formulation. 
Our approach implements this as a functional disentanglement within a two-stage training, decomposing the 3D Gaussians' feature representation into simpler, more manageable sub-problems as illustrated in Fig.~\ref{fig:main}c.
Our two-stage pipeline is intentionally designed to facilitate this separation, even when supervised by the same ground truth images. 
We begin by learning a \emph{geometry-driven base feature} in the first stage. 
By intentionally constraining the input of base network to local, first-order geometric information (unwrapped mesh surface normals), we guide it to learn a stable and low-frequency foundation that represents a robust average appearance. 
These features are predicted in UV-space and mapped onto the Gaussians bound on the driving mesh as their color and opacity attributes.
This results in a coarse but stable representation, where each Gaussian carries base-level appearance aligned to the underlying geometry.

Once the stable base is established, a second network conditioned on both surface normals and FLAME parameters (serving as high-level descriptors) predicts the personalized residual features.
These residuals, also defined in UV-space, are combined with the base feature by dynamic appearance fusion accounting for geometric deformations of each Gaussian.
This separation and fusion enable the model to focus its capacity on capturing high-frequency, non-linear details (e.g., wrinkles) that the base model alone cannot represent.

Our contributions can be summarized as follows:
	1.	We propose DipGuava, a first 3DGS-based framework explicitly disentangling facial appearance into geometry-driven base and personalized residual, significantly reducing learning ambiguity.
	2.	We introduce a UV-based feature representation that enables dynamic appearance fusion, ensuring the effective integration of disentangled features from each stage.
	3.	Extensive evaluations demonstrate that DipGuava significantly outperforms existing approaches achieving superior expressiveness and photorealism.

\section{2. Related Work}
\label{sec:relatedwork}

Creating 3D head avatars from monocular video remains a central challenge in vision and graphics. 
A widely used strategy builds upon 3DMMs~\cite{paysan20093d, gerig2018morphable, li2017learning}, which provide consistent mesh topology and interpretable control over shape and expression parameters. 
While effective for capturing coarse facial geometry, 3DMMs inherently lack the capacity to represent fine-grained details and non-linear skin deformations.

To overcome these limitations, recent works combine 3DMMs with neural rendering techniques, including SDFs~\cite{park2019deepsdf, wang2021prior, zheng2022sdf}, triplanes~\cite{xu2023avatarmav, ma2023otavatar}, and NeRFs~\cite{hong2022headnerf, zhuang2022mofanerf, athar2022rignerf, yao2022dfa, athar2023flame}. A common approach is to learn mappings from a canonical space (neutral expression) to posed and expressive faces. INSTA~\cite{zielonka2023instant} leverages dynamic implicit fields with FLAME guidance, IMAvatar~\cite{zheng2022avatar} uses blendshape-driven SDFs, and point-based methods like PointAvatar~\cite{zheng2023pointavatar} and GPAvatar~\cite{chu2024gpavatar} employ expression-conditioned point representations for detailed modeling.

The emergence of 3DGS~\cite{kerbl20233d} has enabled new approaches that use Gaussian points defined in a geometrically explicit form for rendering.
Methods like GaussianAvatars~\cite{qian2024gaussianavatars} and SplattingAvatar~\cite{shao2024splattingavatar} define Gaussians relative to mesh surfaces or points, animated via FLAME-driven LBS. 
Additionally, PSAvatar~\cite{zhao2024psavatar} aims to represent components outside the face, such as glasses and hair, by geometrically assigning points. 
Gaussian blendshape models were also proposed~\cite{ma20243d, li2025rgbavatar}, where Gaussian features are conditioned on FLAME expression and pose parameters to capture facial dynamics.
FATE~\cite{zhang2025fate} bakes dynamic appearance into Gaussian texture maps, enabling stylization and unseen view regularization.
Several recent works enhance generalization by leveraging Gaussian priors trained on diverse subjects.
GEM~\cite{zielonka2025gaussian} builds a linear basis for lightweight distillation, while HeadGAP~\cite{zheng2024headgap}, SynShot~\cite{zielonka2025synthetic}, and SEGA~\cite{guo2025sega} fine-tune subject-specific models for stable animation.
While successful in capturing geometry and motion, these methods still limit their expressiveness to that of the driving 3DMMs. 

To address this limited detail expressiveness, several methods have proposed modeling \textit{dynamic} changes in Gaussian point features based on expression~\cite{xiang2024flashavatar, xu2024gaussian, chen2024monogaussianavatar}. 
For instance, FlashAvatar~\cite{xiang2024flashavatar} defines geometric deformation in UV space to enhance training efficiency, while MonoGaussianAvatar~\cite{chen2024monogaussianavatar} models expression-conditioned 3D Gaussian deformation fields for monocular reconstruction through per-Gaussian point feature representation. 
However, a key challenge remains in the complexity of mapping from the canonical space to the final expressed appearance, making it difficult for the model to disentangle geometry-driven cues from independently moving, identity-specific details such as subtle expressions and wrinkles.

In contrast, we propose \textbf{DipGuava}, a two-stage framework that \textit{explicitly} disentangles facial appearance into geometry-driven base and personalized residual. 
Rather than jointly predicting both components, we first build a stable base feature and then separately predict personalized residuals.
This separation reduces learning ambiguity and enables the network to capture fine-grained, identity-specific variations even under monocular supervision.

\section{3. Preliminaries} \label{sec:pre}
In our work, a 3D head avatar is composed of multiple Gaussian primitives, each modeled as a spatially localized ellipsoid that captures local geometry and color.
Each Gaussian comprises appearance features (color $c \in \mathbb{R}^3$ and opacity $o \in \mathbb{R}$) and geometric features (position $\mu \in \mathbb{R}^3$, scale $s \in \mathbb{R}^3$, and rotation $r \in \mathbb{R}^4$), which we refer to throughout this paper.

\subsection{3.1 3D Gaussian Binding on Driving Mesh}
Our approach animates facial expressions by adaptively deforming 3D Gaussians bound to the driving FLAME mesh using the binding mechanism in GaussianAvatars~\cite{qian2024gaussianavatars}. 
This binding enables the Gaussians to first move with the facial mesh, capturing the linear movements defined by the 3DMM and to be deformed in local coordinates.
The local space is defined by the triangle's centroid $T$, with the rotation matrix $R$ capturing the triangle's orientation and a scaling factor $k$ representing the mean edge length. 
Each Gaussian is initialized with position $\mu_{l}$ at the origin (each triangle's centroid), rotation $r_{l}$ as an identity matrix, and scale $s_{l}$ as a unit vector. 
These local geometric features are transformed into global space before rendering by:
\begin{equation}
    \begin{aligned}
        r = R r_{l}, \quad
        \mu = k R \mu_{l} + T, \quad
        s = k s_{l}.
    \end{aligned}
\end{equation}

\subsection{3.2 Adaptive UV Feature Sampling}
To establish a shared representation for feature prediction and fusion, we define Gaussian features in the UV space of the FLAME mesh.
For Gaussians not lying directly on the mesh surface, we compute UV coordinates via barycentric interpolation over their associated triangles:
\begin{equation}
p_{\mu_{l}} = \mathbf{UV}(\mu_{l}), \quad \mathbf{x} = \text{Sample}(\mathcal{X}, p_{\mu_{l}}),
\end{equation}
where $\mathcal{X}$ is the UV feature map and $\mathbf{x}$ is the sampled feature.
Additional details are provided in the supplementary material.

\section{4. Method}
\label{sec:method}

\begin{figure*}[t]
\centering
\includegraphics[width=0.99\linewidth]{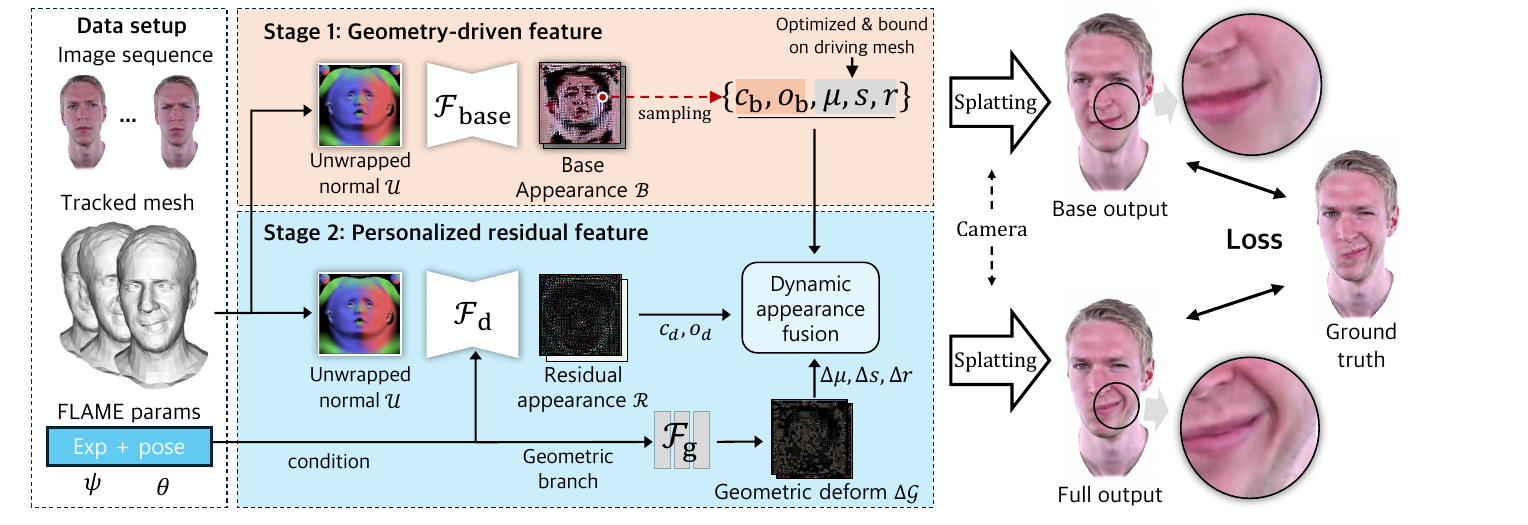}
\caption{
\textbf{Overview of DipGuava}. 
Stage 1 optimizes a geometry-driven base appearance (overall color and opacity from mesh surface normals). 
Stage 2 predicts residual features to capture facial details beyond the base appearance. 
Dynamic appearance fusion combines these residuals with the base appearance and geometric deformations.
}
\label{fig:framework}
\end{figure*}

The overall pipeline of DipGuava is illustrated in Figure~\ref{fig:framework}.
To model fine, identity-specific dynamics beyond the expressive range of 3DMMs, we adopt a two-stage training scheme that explicitly disentangles geometry-driven base appearance from personalized residuals.
This separation allows the model to first learn a stable, interpretable foundation, then refine it with residual details, enabling accurate modeling of non-linear deformations and subtle expressions.

\subsection{4.1 Disentanglement of Base and Residual Features}
\label{sec:disentanglement}

\subsubsection{Geometry-driven Base Appearance Training}
We begin by establishing a \textit{geometry-driven base appearance} that represents the global facial structure and attributes aligned with the driving expression and pose. 
To achieve this, we rasterize the FLAME model’s normals into UV space, producing a geometry-aware UV normal map $\mathcal{U}$.
This is done via barycentric interpolation of vertex normals oriented with respect to the camera, allowing the UV map to accurately capture local surface orientation and curvature.
This unwrapped normal map $\mathcal{U}$ is then fed into the \textbf{Base appearance network} $\mathcal{F}_{\text{base}}$, a U-Net architecture, which learns to predict the base appearance map $\mathcal{B}$.

During stage 1, for each 3D Gaussian primitive $i$, we jointly optimize its geometric features $\mu_l, s_l, r_l$. 
To determine the base color and opacity for this Gaussian, we sample the base appearance map $\mathcal{B}$ at the corresponding UV coordinates $p_{\mu_{l}}$, obtained by projecting the Gaussian's position onto the UV space: $p_{\mu_{l}} = \mathbf{UV}(\mu_l)$. 
The base color $c_{\text{b}}$ and opacity $o_{\text{b}}$ are then sampled as $(c_{\text{b}}, o_{\text{b}}) = \text{Sample}(\mathcal{B}, p_{\mu_{l}}).$
Combined with the geometric attributes $\mu, s, r$ which are converted from their local forms $\mu_l, s_l, r_l$ into global space as described in Sec. 3.1, the full Gaussian representation at the first stage is:
\begin{equation}
\mathit{G} = \{c_{\text{b}}, o_{\text{b}}, \mu, s, r\}.
\end{equation}


\subsubsection{Personalized Residual Appearance Prediction}
The second stage focuses on modeling non-linear facial features, including high-frequency details like fine wrinkles and hair strands, which are not fully captured by the geometry-driven base feature. 
To this end, residual features for both appearance and geometry are predicted, aiming to reduce the discrepancy between the stage 1 output and the ground truth image.

With the geometric positions of the Gaussian points optimized in Stage 1 and the Base Appearance Network $\mathcal{F}_{\text{base}}$ frozen, we train a \textbf{Dynamic appearance network} $\mathcal{F}_{\text{d}}$. 
This network takes the same normal-based UV map $\mathcal{U}$ as input, along with a condition vector comprising FLAME expression parameters $\psi$ and pose parameters $\theta$. 
This dual-input design is intentional: while the normal map provides dense, low-level geometric detail, the FLAME parameters act as a compact, high-level semantic guide for the expression. 
This condition vector is concatenated at the U-Net's bottleneck to modulate the network's output toward the desired expression and pose.
The network predicts a residual appearance map $\mathcal{R}$ as:
\begin{equation}
\mathcal{R} = \mathcal{F}_{\text{d}}(\mathcal{U}, \psi, \theta).
\end{equation}
In parallel, we employ a geometric MLP, $\mathcal{F}_{\text{g}}$, which takes the same condition vector (FLAME expression parameters $\psi$ and pose parameters $\theta$) as input and predicts a geometric deformation map $\Delta\mathcal{G}$ in UV space:
\begin{equation}
\Delta\mathcal{G} = \mathcal{F}_{\text{g}}(\psi, \theta).
\end{equation}
This design enables the prediction of both appearance and geometric residuals conditioned on shared FLAME-based semantic parameters, thus maintaining a clear disentanglement from the base geometry-driven representation established in the first stage.

\begin{figure}[t]
\centering
\includegraphics[width=0.99\linewidth]{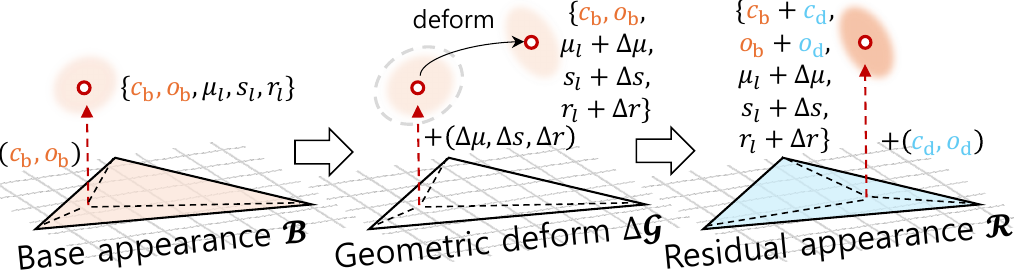}
\caption{\textbf{Dynamic appearance fusion.} To ensure alignment, personalized residual appearance sampled from UV space is combined with the base appearance \emph{after} the geometric deformation.}
\label{fig:fusion}
\end{figure}

\subsection{4.2 Dynamic Appearance Fusion}
Simply adding residual features sampled from the original location of a Gaussian point fails to account for deformation-induced shifts. 
As a result, these residual features can become misaligned with the intended facial region, leading to appearance artifacts or blurred details.
To address this, we introduce a geometry-aware residual fusion strategy that resamples dynamic appearance features at the updated UV coordinates \emph{after} the geometric deformation. 
This ensures that residual appearance features are sampled from the UV location corresponding to the deformed Gaussian’s semantic context. 
This design avoids erroneous blending and allows for more precise correction of expression-specific variations.

During the fusion process, the Gaussian features are obtained by combining geometry-driven and residual features, as depicted in Figure~\ref{fig:fusion}. 
Each Gaussian’s optimized position $\mu_{l}$ from the first stage is used to sample its base appearance $(c_b, o_b)$.
Next, residual geometric deltas in local coordinates $(\Delta\mu, \Delta s, \Delta r)$ are sampled from $\Delta\mathcal{G}$ and applied as:
\begin{equation}
\mu'_{l} = \mu_{l} + \Delta\mu, \quad s'_{l} = s_{l} + \Delta s, \quad r'_{l} = r_{l} + \Delta r.
\end{equation}
At the updated UV position, personalized residuals $(c_\text{d}, o_\text{d})$ are then sampled from $\mathcal{R}$ and added to the base:
\begin{equation}
p_{\mu'_{l}} = \mathbf{UV}(\mu'_{l}), \quad (c_\text{d}, o_\text{d}) = \text{Sample}(\mathcal{R}, p_{\mu'_{l}}),
\end{equation}
\begin{equation}
c = c_\text{b} + c_\text{d}, \quad o = o_\text{b} + o_\text{d}.
\end{equation}
After transforming the geometry into global space, the final representation becomes:
\begin{equation}
\mathit{G} = \{c, o, \mu', s', r'\}.
\end{equation}

\subsection{4.3 Gaussian Splatting Rendering}
We render the output image using 3D Gaussian splatting as introduced by Kerbl~\emph{et al.}~\cite{kerbl20233d}. 
In our framework, as shown in Fig.~\ref{fig:framework}, this rendering process is applied at both training stages. 
During stage 1, only the geometry-driven base features are used for rendering. 
In stage 2, we render the fused features that incorporate residual appearance and residual geometric deformation.

\subsection{4.4 Training Objectives}
\label{sec:training}
Both Stage 1 and Stage 2 are supervised with the same GT images.
In Stage 1, the loss is computed from the rendered output using geometry-driven base features.
In Stage 2, it is computed on the final fused image after dynamic appearance fusion.
The total loss includes photometric, perceptual, and geometric regularization terms.
\begin{equation}
\mathcal{L}_{\text{total}} = \mathcal{L}_{\text{pho}} + \lambda_{\text{lpips}} \mathcal{L}_{\text{lpips}} + \lambda_{\text{xyz}} \mathcal{L}_{\text{xyz}} +  \lambda_{\text{scale}} \mathcal{L}_{\text{scale}}.
\end{equation}
We use a combination of $L_1$ and SSIM losses:
\begin{equation}
\mathcal{L}_{\text{pho}} =  \lambda_{\text{L1}} | I - I_{\text{gt}} |_1 +  (1-\lambda_{\text{L1}})\left(1 - \text{SSIM}(I, I_{\text{gt}})\right).
\end{equation}
To encourage perceptual similarity, we include an LPIPS loss~\cite{zhang2018unreasonable}:
\begin{equation}
\mathcal{L}_{\text{lpips}} =  | \phi(I) - \phi(I_{\text{gt}}) |_2^2.
\end{equation}
where $\phi$ denotes a VGG-based feature extractor.
Following prior work~\cite{qian2024gaussianavatars}, we regularize both the original and deformed 3D Gaussian primitives by penalizing out-of-bounds positions and overly large scales with thresholds $\epsilon_{\text{xyz}}$ and $\epsilon_{\text{scale}}$. 
Specifically, the loss is applied to both the stage-1 geometry $(\mu_l, s_l)$ and the predicted stage-2 geometry $(\mu'_l, s'_l)$.

\begin{equation}
\mathcal{L}_{\text{xyz}} = 
\sum_{\tilde{\mu} \in \{\mu_{l}, \mu'_{l}\}} \left\| \max\left(\|\tilde{\mu}\|_2 - \epsilon_{\text{xyz}},\ 0\right) \right\|_2,
\end{equation}
\begin{equation}
\mathcal{L}_{\text{scale}} = 
\sum_{\tilde{s} \in \{s_{l}, s'_{l}\}} \left\| \max\left( \exp(\tilde{s}) - \epsilon_{\text{scale}},\ 0 \right) \right\|_2.
\end{equation}
\section{5. Experiments}
\label{sec:experiments}

\subsubsection{Datasets}
Our method uses a monocular video of a single subject as input. 
We use videos from NHA~\cite{grassal2022neural}, NerFace~\cite{gafni2021dynamic}, PointAvatar~\cite{zheng2023pointavatar}, and INSTA~\cite{zielonka2023instant} for 11 subjects. 
Additionally, we captured 7 more videos under diverse conditions, including both controlled studio settings and uncontrolled scenarios using a mobile phone. 
In total, 18 videos with various environments were used in our experiments. 
Each video ranges 1–3 minutes (512x512 resolution). 
For a fair comparison, we use the same tracking results (FLAME parameters and meshes with teeth and vertex deformations around the hairline contour), masked images~\cite{yu2021bisenet}, face-camera rotation from VHAP~\cite{qian2024vhap} across all sequences, and we set the last 30\% of frames from each subject as the testing set.

\subsubsection{Implementation Details}
We train the model in two stages, each for 60k iterations.
In stage 1, we jointly optimize the base geometry and $\mathcal{F}_\text{base}$ using a learning rate of $1 \times 10^{-4}$. 
Densification, cloning, and splitting are performed every 500 iterations, with Gaussians having opacity below 0.05 pruned.
After stage 1, both the optimized geometry and the weights of $\mathcal{F}_\text{base}$ are frozen. 
In stage 2, the residual appearance network and geometric MLP are trained with a learning rate of $1 \times 10^{-5}$.
We use the following loss weights: $\lambda_{\text{L1}}=0.8, \lambda_{\text{LPIPS}} = 0.1, \lambda_{\text{xyz}} = 0.01, \lambda_{\text{scaling}} = 1.0,$ and geometric threshold in local scale: $\epsilon_{\text{xyz}} = 2.0, \epsilon_{\text{scale}} = 0.6.$
All UV maps have a resolution of 128x128.

\subsection{5.1 Comparison with SOTA Methods}
To evaluate the effectiveness of proposed method, we conducted comprehensive experiments comparing DipGuava with a wide range of SOTA 3D head avatar approaches. 
These include PointAvatar (PA)~\cite{zheng2023pointavatar}, INSTA~\cite{zielonka2023instant}, GaussianAvatars (GA)~\cite{qian2024gaussianavatars}, FlashAvatar (FA)~\cite{xiang2024flashavatar}, SplattingAvatar (SA)~\cite{shao2024splattingavatar}, MonoGaussianAvatar (MGA)~\cite{chen2024monogaussianavatar}, and FATE~\cite{zhang2025fate}.
Among these, we include qualitative comparisons with methods that produce reasonably comparable outputs. 
Please refer to the supplementary material for visual comparisons against all methods and per-subject results.

\begin{table}[t]
    \centering
    \resizebox{0.47\textwidth}{!}{
      \begin{tabular}{ccccc}
        \hline\hline
        Method & L1 $\downarrow$ & LPIPS $\downarrow$ & SSIM $\uparrow$ & PSNR $\uparrow$ \\
        \hline\hline
        PointAvatar & 0.016 & 0.056 & 0.923 & 25.21 \\
        INSTA & 0.016 & 0.108 & 0.907 & 24.73 \\
        GaussianAvatars & \underline{0.012} & 0.066 & 0.944 & 28.36 \\
        FlashAvatar & 0.018 & 0.071 & 0.918 & 25.72 \\
        SplattingAvatar & 0.019 & 0.096 & 0.917 & 25.61 \\
        MonoGaussianAvatar & 0.014 & 0.053 & 0.939 & 26.37 \\
        FATE & 0.012 & \underline{0.045} & \underline{0.945} & \underline{28.38} \\ \hline
        \textbf{Ours} & \textbf{0.009} & \textbf{0.042} & \textbf{0.957} & \textbf{30.14} \\
        \hline\hline
      \end{tabular}
    }
    \caption{Quantitative comparison with SOTA methods. \textbf{Best} and \underline{second-best} results highlighted.}
    \label{tab:benchmark}
\end{table}

\begin{figure}
    \centering
    \includegraphics[width=0.95\linewidth]{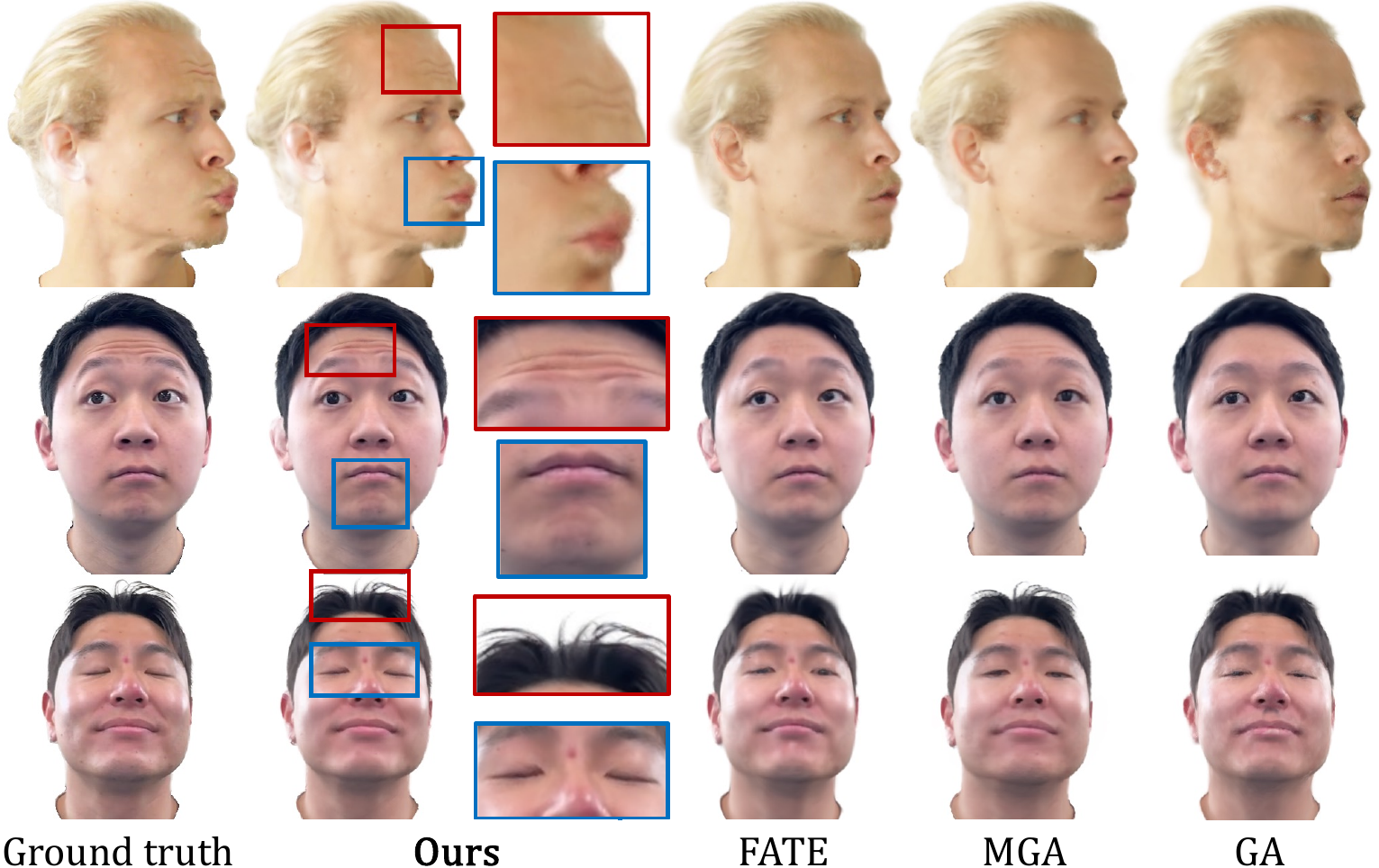}
    \caption{\textbf{Qualitative comparison in self-driven animation.} Our method generates outperforming results with details such as wrinkles, eye blinks, and lip movement.}
    \label{fig:selfdriven}
\end{figure}

\subsubsection{Training Time and Inference Speed}
Under identical settings (RTX A6000), our method converges in 70 minutes, which is comparable to existing approaches (FATE: 45m, MGA: 9h, SA: 44m, FA: 20m, GA: 40m, INSTA: 60m, PA: 7h), while achieving superior performance in capturing subtle expressions and details. 
At inference time, our full model runs at 88 FPS (512×512 resolution), enabling real-time applications.

\subsubsection{Quantitative Performance Comparison}
For quantitative evaluation, we use L1 distance, LPIPS~\cite{zhang2018unreasonable}, SSIM~\cite{wang2004image}, and PSNR.
Table~\ref{tab:benchmark} summarizes the average performance across all identities in our benchmark.
DipGuava consistently outperforms all baselines across all metrics, clearly demonstrating its effectiveness under identical training conditions.
The performance gap arises from the fact that only our method accurately reconstructs fine-grained, identity-specific attributes. 
While previous approaches may look acceptable without ground truth, they fail to capture the personalized details critical for realism. 
This distinction is evident in the following qualitative results.

\begin{figure}[t]
  \centering
  \includegraphics[width=0.99\linewidth]{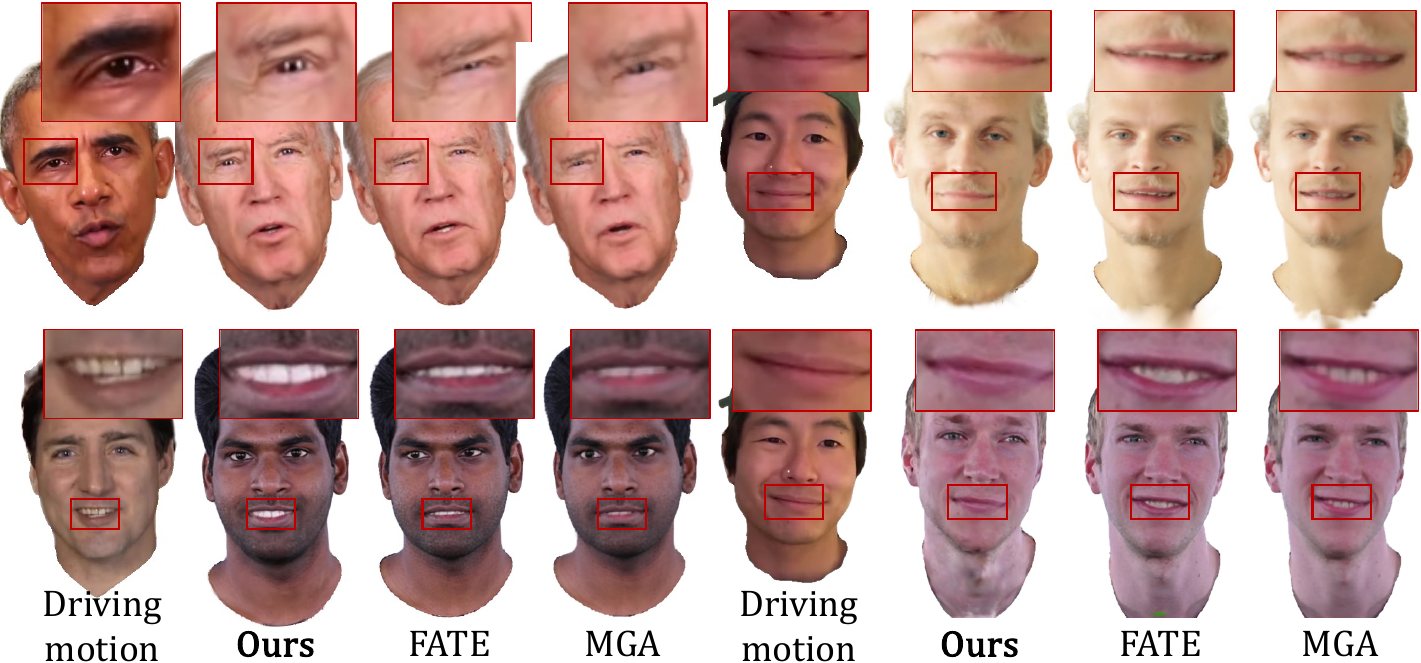}
  \caption{\textbf{Qualitative comparison in cross-id reenactment.} The proposed method preserves both facial structure and appearance with high fidelity, while accurately following subtle expressions in the driving motion.}
  \label{fig:retarget}
\end{figure}

\subsubsection{Self-driven Re-animation}
Figure~\ref{fig:selfdriven} presents a qualitative comparison of DipGuava with prior state-of-the-art methods. 
Overall, DipGuava produces sharper and more vivid outputs while faithfully reconstructing personalized appearance and expression-driven geometry.

In both the first and second subjects, DipGuava is the only method that accurately reconstructs \textit{forehead wrinkles}, which are entirely missing or oversmoothed in other methods. 
This highlights DipGuava’s strength in preserving identity-specific high-frequency features.
For the first subject, DipGuava captures subtle lip protrusion under a side-profile view, which others fail to represent.
In the second subject, the jawline and eyelids follow expression-induced shape changes with clear articulation, especially in wide-eye expressions with upward gaze.
The third subject further demonstrates DipGuava’s robustness in reconstructing closed-eye expressions without introducing artifacts, unlike prior works that either hallucinate eyes or leave them open. Additionally, fine structures such as raised bangs are reconstructed with greater accuracy.

\subsubsection{Cross-identity Reenactment}
In Fig.~\ref{fig:retarget}, we evaluate cross-identity reenactment performance, comparing our method with recent models which show reasonable results in this task.
A key distinction lies in the fidelity of expression transfer. 
DipGuava more accurately conveys subtle motions, such as slight smiles and eye blinks, while preserving identity.
In contrast, both FATE and MGA often exhibit entangled modeling of lips and teeth, frequently revealing teeth even in closed-mouth expressions, and rendering eye movements unnaturally. 
Notably, DipGuava consistently reproduces gaze direction and eyelid motion with higher precision, even in unseen, subtle expressions such as closed-mouth smiles or eyelid behavior.

\begin{table*}
\centering
\resizebox{2.0\columnwidth}{!}{%
\begin{tabular}{cccccccccccc}
\hline\hline
 & \multicolumn{2}{c}{Training strategy}
 & \multicolumn{3}{c}{Model design}
 & \multicolumn{5}{c}{Model component}
 & \textbf{\textit{Full}} \\ 
\cmidrule(lr){2-3} \cmidrule(lr){4-6} \cmidrule(lr){7-11}

Metric 
& \textit{Joint} & \textit{LowRes} 
& \textit{OptBase} & \textit{FixedUV} & \textit{MLP} 
& \textit{only} $\mathcal{B}$  & \textit{only} $\mathcal{R}$  & \textit{w/o} $\Delta\mathcal{G}$ & \textit{w/o} $\mathcal{R}$ & \textit{w/o DAF} 
& $\mathcal{B}$+$\mathcal{R}$+$\Delta\mathcal{G}$ \\ 
\cmidrule(lr){1-12}
L1 $\downarrow$
& 0.013 & 0.011 & 0.012 & 0.010 & 0.010 
& 0.0099 & 0.0130 & 0.0093 & 0.0092 & \underline{0.0091} 
& \textbf{0.0090} \\
LPIPS $\downarrow$
& 0.0491 & 0.0949 & 0.0492 & 0.0453 & 0.0516 
& 0.0433 & 0.0955 & 0.0440 & 0.0434 & \textbf{0.0410} 
& \underline{0.0417} \\
SSIM $\uparrow$
& 0.936 & 0.949 & 0.944 & 0.952 & 0.938 
& 0.9538 & 0.9357 & 0.9547 & 0.9553 & \underline{0.9554} 
& \textbf{0.9570} \\
PSNR $\uparrow$
& 29.02 & 29.42 & 28.70 & 29.53 & 28.32 
& 29.54 & 28.25 & 29.94 & \underline{29.95} & 29.86 
& \textbf{30.14} \\
\hline\hline
\end{tabular}
}
\caption{Quantitative result of training strategies, model designs, and ablations. \textbf{Best} and \underline{second-best} results are highlighted.}

\label{tab:ablation_full}
\end{table*}
\subsection{5.2 Analysis of Training and Model Design Choices}
Since our method integrates a two-stage training pipeline with multiple components, we conduct a comprehensive analysis to validate the effectiveness of each design choice as shown in Tab.~\ref{tab:ablation_full}.

\subsubsection{Effect of Training Strategy}
We compare two variants to evaluate convergence and detail fidelity.
\textbf{\textit{Joint}} fine-tunes both $\mathcal{F}\text{base}$ and residual features with a reduced learning rate for the base.
\textbf{\textit{LowRes}} trains $\mathcal{F}\text{base}$ at $256^2$ resolution, then switches to $512^2$ for residual refinement.
Our staged strategy achieves superior convergence and recovers high-frequency details more effectively.

\subsubsection{Effect of Model Design Choice}
We assess key architectural components via ablations.
\textbf{\textit{OptBase}} removes surface normal input and optimizes appearances, degrading generalization to unseen expressions.
\textbf{\textit{FixedUV}} disables adaptive UV sampling, limiting expression-specific details.
\textbf{\textit{MLP}} replaces the UNet backbone with an MLP-based model, resulting in worse performance, especially in high-frequency regions. This confirms the benefit of spatially-aware convolution in UV space.

\begin{figure}[t]
    \centering
    \includegraphics[width=0.95\linewidth]{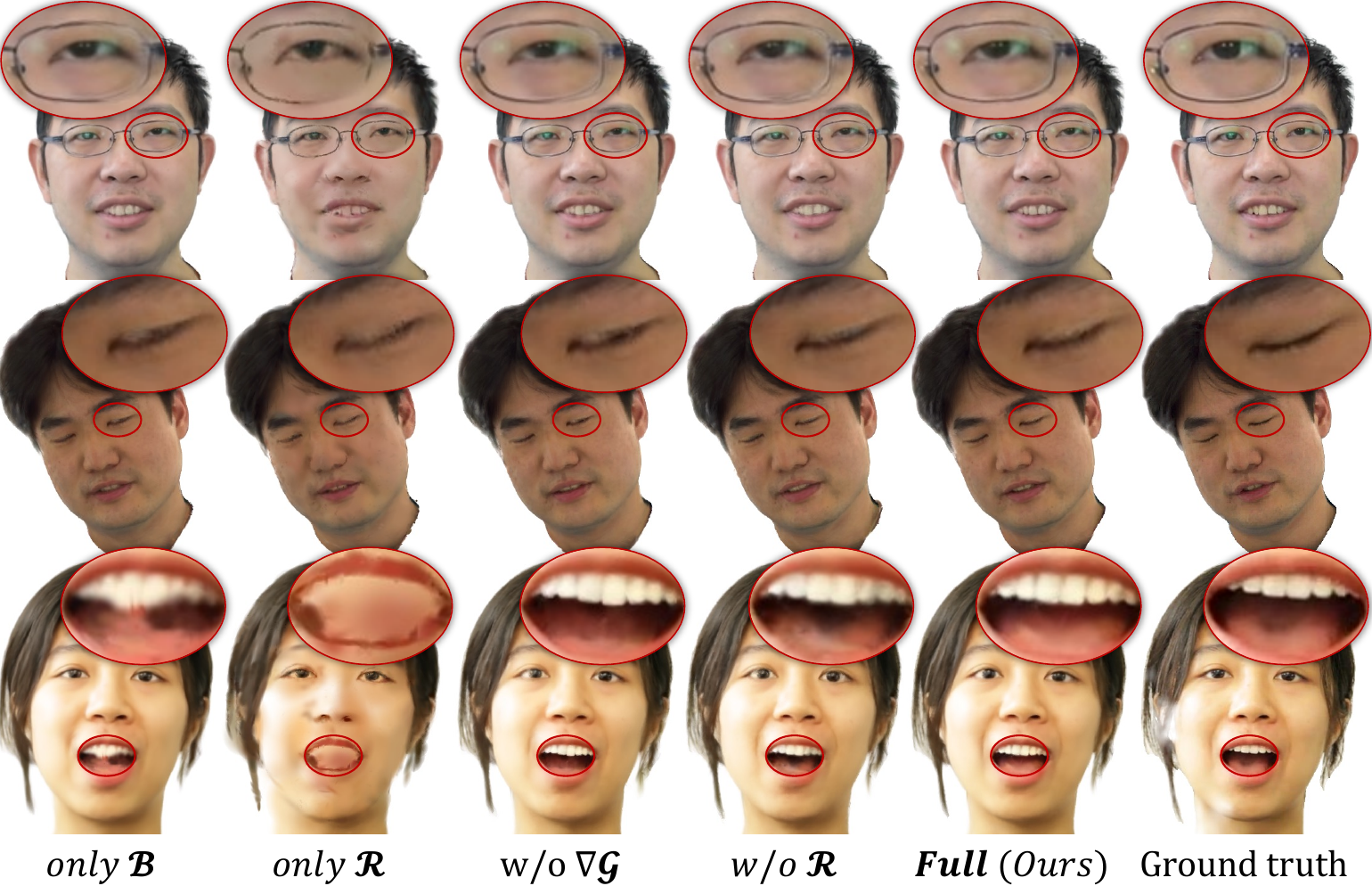}
\caption{\textbf{Impact of individual components.} The full model preserves structure and fine-scale details most effectively.}
    \label{fig:ablation}
\end{figure}

\subsection{5.3 Ablations for Model Components}

We conduct detailed ablations to evaluate the contribution of each component in our framework as shown in Tab.~\ref{tab:ablation_full} and Fig.~\ref{fig:ablation}.
Here, ``\textbf{\textit{only} $\mathcal{B}$}'' and ``\textbf{\textit{only} $\mathcal{R}$}'' represent single-stage models that predict appearance using only the base or residual branch, respectively.
For the two-stage setup, we evaluate ``\textbf{\textit{w/o} $\mathcal{R}$}'', ``\textbf{\textit{w/o} $\Delta\mathcal{G}$}'', and disabling dynamic appearance fusion by simply summing $\mathcal{B}$ and $\mathcal{R}$ (``\textbf{\textit{w/o DAF}}'').

\subsubsection{Effectiveness of Two-stage Feature Disentanglement}
Fig.~\ref{fig:ablation} shows that while the base appearance captures the overall facial layout, personalized wrinkles and high-frequency attributes such as eyeglass frames are preserved only when residual features are incorporated.
Relying solely on $\mathcal{R}$ leads to significant performance drops across all metrics and introduces noticeable artifacts like blurred fine structures that 3DMMs struggle to represent.
Our full two-stage model consistently outperforms all ablations, demonstrating the benefit of explicit feature disentangling.

\subsubsection{Complementary Roles of Residual Appearance and Geometric Deformation}
Within our two-stage framework, the residual appearance and geometric deformation play distinct yet complementary roles. 
As shown in Table~\ref{tab:ablation_full} and Fig.~\ref{fig:ablation}, removing either residual component leads to a consistent drop in performance, indicating that relying on either color correction or geometric adjustment alone is insufficient to capture subtle motion and appearance. 
For example, accurate eyeglass reconstruction requires both color refinement from the residual appearance and precise geometric adjustments to align the Gaussians with the frame's shape.

\begin{figure}[t]
    \centering
    \includegraphics[width=0.95\linewidth]{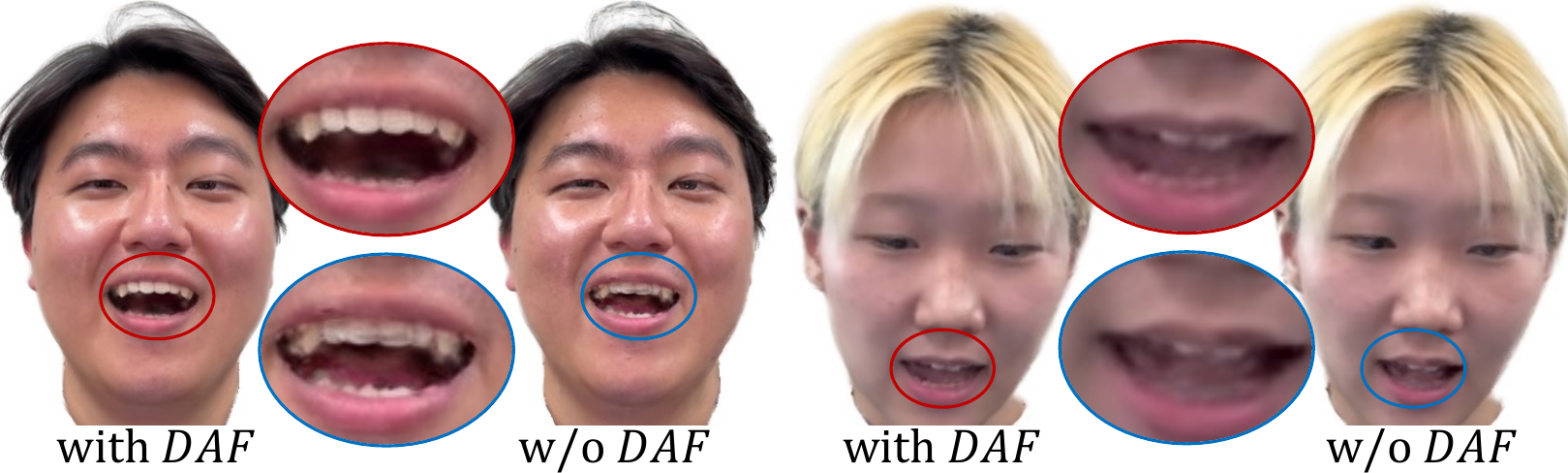}
\caption{Applying \textbf{dynamic appearance fusion} yields sharper and less noisy results in regions with rapid color changes, such as inside the mouth.}
    \label{fig:daf}
\end{figure}

\subsubsection{Importance of Dynamic Appearance Fusion}
Interestingly, while \textit{w/o DAF} achieves comparable performance in LPIPS compared to proposed full method, Fig.~\ref{fig:daf} reveals that omitting dynamic appearance fusion leads to noise and artifacts in semantically complex regions like the mouth interior.
In other words, applying residual appearance features from pre-deformation UV coordinates results in mismatched textures.
In contrast, our geometry-aware fusion re-samples residuals \textit{after} deformation, ensuring proper alignment and enabling the model to capture non-linear variations more effectively.

\begin{figure}[t]
  \centering
  \includegraphics[width=0.95\linewidth]{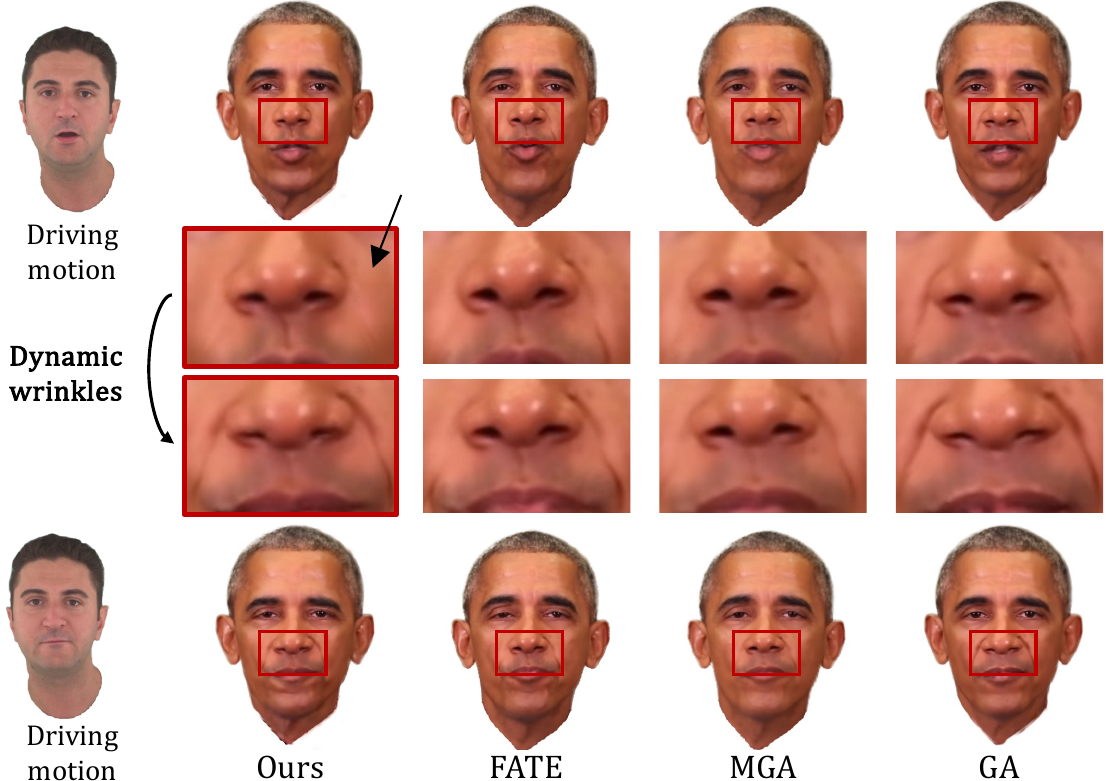}
\caption{\textbf{Expression-aware dynamic wrinkles.} Unlike other methods that model wrinkles as static color textures, our method adaptively captures them in response to facial motion and geometry.}
  \label{fig:wrinkle}
\end{figure}
\begin{figure}[t]
    \centering
    \includegraphics[width=0.95\linewidth]{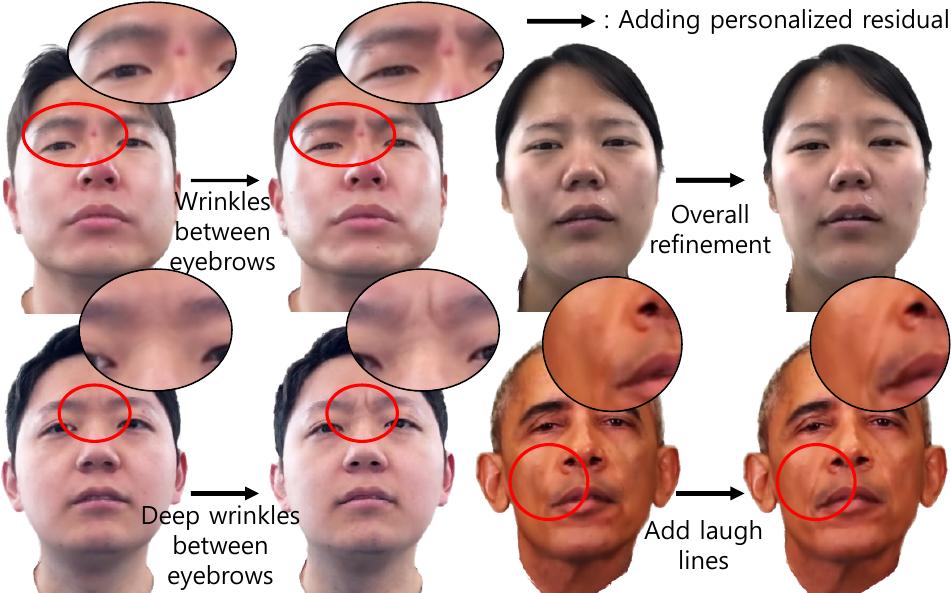}
\caption{\textbf{Personalized residuals from the same driving motion.} Residual features inject subject-specific details, enabling personalized expressions from identical motions.}
    \label{fig:personalization}
\end{figure}

\subsection{5.4 Analysis of Personalized Residual Feature}
\subsubsection{Dynamic Wrinkle Modeling}
Fig.~\ref{fig:wrinkle} further illustrates the dynamic wrinkle modeling ability of DipGuava. Unlike other methods that render wrinkles as static textures unaffected by expression, our method generates expression-dependent wrinkles in a natural and temporally coherent manner. For example, in the upper row, wrinkles around the mouth are smoothed as the lips move forward, while in the lower row, new wrinkles emerge in response to the expression, faithfully reflecting the underlying skin deformation.

\subsubsection{Id-specific Attribute in Personalized Residuals}
Beyond simply enhancing detail, our method models personalized residuals in an identity-specific manner, as shown in Fig.~\ref{fig:personalization}.
While the base model captures the shared expression across subjects, the residual features add unique, subject-specific attributes, effectively disentangling individual characteristic from the shared motion.
For example, the model accurately synthesizes localized wrinkles for subjects who naturally exhibit them, while preserving smooth skin for others without introducing hallucinated details.
This demonstrates that our method learns not only to refine appearance, but also to model personalized, identity-consistent dynamics.
\section{6. Conclusion and Limitations}
\label{sec:conclusion}
We presented DipGuava, a novel framework that creates high-fidelity, animatable 3D head avatars from monocular video. 
Our key contribution is a two-stage approach that explicitly disentangles the complex facial representation into a geometry-driven base and a personalized residual. 
This decomposition resolves the learning ambiguity of prior methods, enabling the targeted modeling of high-frequency, subject-specific details. 
Finally, combined with a dynamic fusion mechanism, our method achieves fine-grained reconstruction of personalized facial features.
Extensive experiments across multiple benchmarks demonstrate that DipGuava outperforms previous methods, offering a practical solution for real-world 3D head avatar applications.
However, our method shares a inherent limitation among 3DMM-driven approaches such as contrained robustness by the fidelity of the 3DMM tracking and the model's capacity to represent poorly sampled areas, such as intra-oral regions or extreme expressions.
Nevertheless, we argue that our disentangled architecture is highly scalable. 
As tracking fidelity improves, our base model provides a cleaner foundation, enabling the residual model to learn finer details and further widen the performance gap over holistic approaches.

\section*{Acknowledgements}
This research was supported by Culture, Sports and Tourism R\&D Program through the Korea Creative Content Agency grant funded by the Ministry of Culture, Sports and Tourism in 2024 (RS-2024-00398413, Contribution Rate: 90\%), the National Research Foundation of Korea(NRF) grant funded by the Korea government(MSIT)(RS-2025-02216328), and the Yonsei Signature Research Cluster Program of 2025 (2025-22-0013).

\bibliography{aaai2026}

@article{mildenhall2021nerf,
  title={Nerf: Representing scenes as neural radiance fields for view synthesis},
  author={Mildenhall, Ben and Srinivasan, Pratul P and Tancik, Matthew and Barron, Jonathan T and Ramamoorthi, Ravi and Ng, Ren},
  journal={Communications of the ACM},
  volume={65},
  number={1},
  pages={99--106},
  year={2021},
  publisher={ACM New York, NY, USA}
}

@inproceedings{gafni2021dynamic,
  title={Dynamic neural radiance fields for monocular 4d facial avatar reconstruction},
  author={Gafni, Guy and Thies, Justus and Zollhofer, Michael and Nie{\ss}ner, Matthias},
  booktitle={Proceedings of the IEEE/CVF Conference on Computer Vision and Pattern Recognition},
  pages={8649--8658},
  year={2021}
}

@inproceedings{hong2022headnerf,
  title={Headnerf: A real-time nerf-based parametric head model},
  author={Hong, Yang and Peng, Bo and Xiao, Haiyao and Liu, Ligang and Zhang, Juyong},
  booktitle={Proceedings of the IEEE/CVF Conference on Computer Vision and Pattern Recognition},
  pages={20374--20384},
  year={2022}
}

@inproceedings{grassal2022neural,
  title={Neural head avatars from monocular rgb videos},
  author={Grassal, Philip-William and Prinzler, Malte and Leistner, Titus and Rother, Carsten and Nie{\ss}ner, Matthias and Thies, Justus},
  booktitle={Proceedings of the IEEE/CVF Conference on Computer Vision and Pattern Recognition},
  pages={18653--18664},
  year={2022}
}

@inproceedings{zheng2022avatar,
  title={Im avatar: Implicit morphable head avatars from videos},
  author={Zheng, Yufeng and Abrevaya, Victoria Fern{\'a}ndez and B{\"u}hler, Marcel C and Chen, Xu and Black, Michael J and Hilliges, Otmar},
  booktitle={Proceedings of the IEEE/CVF Conference on Computer Vision and Pattern Recognition},
  pages={13545--13555},
  year={2022}
}

@inproceedings{athar2023flame,
  title={Flame-in-nerf: Neural control of radiance fields for free view face animation},
  author={Athar, ShahRukh and Shu, Zhixin and Samaras, Dimitris},
  booktitle={2023 IEEE 17th International Conference on Automatic Face and Gesture Recognition (FG)},
  pages={1--8},
  year={2023},
  organization={IEEE}
}

@inproceedings{zielonka2023instant,
  title={Instant volumetric head avatars},
  author={Zielonka, Wojciech and Bolkart, Timo and Thies, Justus},
  booktitle={Proceedings of the IEEE/CVF Conference on Computer Vision and Pattern Recognition},
  pages={4574--4584},
  year={2023}
}

@article{kerbl20233d,
  title={3D Gaussian Splatting for Real-Time Radiance Field Rendering.},
  author={Kerbl, Bernhard and Kopanas, Georgios and Leimk{\"u}hler, Thomas and Drettakis, George},
  journal={ACM Trans. Graph.},
  volume={42},
  number={4},
  pages={139--1},
  year={2023}
}

@article{kirschstein2023nersemble,
  title={Nersemble: Multi-view radiance field reconstruction of human heads},
  author={Kirschstein, Tobias and Qian, Shenhan and Giebenhain, Simon and Walter, Tim and Nie{\ss}ner, Matthias},
  journal={ACM Transactions on Graphics (TOG)},
  volume={42},
  number={4},
  pages={1--14},
  year={2023},
  publisher={ACM New York, NY, USA}
}

@inproceedings{zheng2023pointavatar,
  title={Pointavatar: Deformable point-based head avatars from videos},
  author={Zheng, Yufeng and Yifan, Wang and Wetzstein, Gordon and Black, Michael J and Hilliges, Otmar},
  booktitle={Proceedings of the IEEE/CVF conference on computer vision and pattern recognition},
  pages={21057--21067},
  year={2023}
}

@inproceedings{qian2024gaussianavatars,
  title={Gaussianavatars: Photorealistic head avatars with rigged 3d gaussians},
  author={Qian, Shenhan and Kirschstein, Tobias and Schoneveld, Liam and Davoli, Davide and Giebenhain, Simon and Nie{\ss}ner, Matthias},
  booktitle={Proceedings of the IEEE/CVF Conference on Computer Vision and Pattern Recognition},
  pages={20299--20309},
  year={2024}
}

@inproceedings{xiang2024flashavatar,
  title={FlashAvatar: High-fidelity Head Avatar with Efficient Gaussian Embedding},
  author={Xiang, Jun and Gao, Xuan and Guo, Yudong and Zhang, Juyong},
  booktitle={Proceedings of the IEEE/CVF Conference on Computer Vision and Pattern Recognition},
  pages={1802--1812},
  year={2024}
}

@inproceedings{chen2024monogaussianavatar,
  title={Monogaussianavatar: Monocular gaussian point-based head avatar},
  author={Chen, Yufan and Wang, Lizhen and Li, Qijing and Xiao, Hongjiang and Zhang, Shengping and Yao, Hongxun and Liu, Yebin},
  booktitle={ACM SIGGRAPH 2024 Conference Papers},
  pages={1--9},
  year={2024}
}

@article{zhao2024psavatar,
  title={Psavatar: A point-based morphable shape model for real-time head avatar creation with 3d gaussian splatting},
  author={Zhao, Zhongyuan and Bao, Zhenyu and Li, Qing and Qiu, Guoping and Liu, Kanglin},
  journal={arXiv preprint arXiv:2401.12900},
  year={2024}
}

@inproceedings{zhang2018unreasonable,
  title={The unreasonable effectiveness of deep features as a perceptual metric},
  author={Zhang, Richard and Isola, Phillip and Efros, Alexei A and Shechtman, Eli and Wang, Oliver},
  booktitle={Proceedings of the IEEE conference on computer vision and pattern recognition},
  pages={586--595},
  year={2018}
}

@article{li2017learning,
  title={Learning a model of facial shape and expression from 4D scans.},
  author={Li, Tianye and Bolkart, Timo and Black, Michael J and Li, Hao and Romero, Javier},
  journal={ACM Trans. Graph.},
  volume={36},
  number={6},
  pages={194--1},
  year={2017}
}

@inproceedings{gerig2018morphable,
  title={Morphable face models-an open framework},
  author={Gerig, Thomas and Morel-Forster, Andreas and Blumer, Clemens and Egger, Bernhard and Luthi, Marcel and Sch{\"o}nborn, Sandro and Vetter, Thomas},
  booktitle={2018 13th IEEE international conference on automatic face \& gesture recognition (FG 2018)},
  pages={75--82},
  year={2018},
  organization={IEEE}
}

@InProceedings{wang2022faceverse,
title={FaceVerse: a Fine-grained and Detail-controllable 3D Face Morphable Model from a Hybrid Dataset},
author={Wang, Lizhen and Chen, Zhiyua and Yu, Tao and Ma, Chenguang and Li, Liang and Liu, Yebin},
booktitle={IEEE Conference on Computer Vision and Pattern Recognition (CVPR2022)},
month={June},
year={2022},
}

@inproceedings{paysan20093d,
  title={A 3D face model for pose and illumination invariant face recognition},
  author={Paysan, Pascal and Knothe, Reinhard and Amberg, Brian and Romdhani, Sami and Vetter, Thomas},
  booktitle={2009 sixth IEEE international conference on advanced video and signal based surveillance},
  pages={296--301},
  year={2009},
  organization={Ieee}
}

@misc{qian2024vhap,
  title={VHAP: Versatile Head Alignment with Adaptive Appearance Priors},
  author={Qian, Shenhan},
  year={2024},
  month={sep},
  doi={10.5281/zenodo.14988309},
  howpublished = {\url{https://github.com/ShenhanQian/VHAP}}
}

@inproceedings{park2019deepsdf,
  title={Deepsdf: Learning continuous signed distance functions for shape representation},
  author={Park, Jeong Joon and Florence, Peter and Straub, Julian and Newcombe, Richard and Lovegrove, Steven},
  booktitle={Proceedings of the IEEE/CVF conference on computer vision and pattern recognition},
  pages={165--174},
  year={2019}
}

@article{wang2021prior,
  title={Prior-guided multi-view 3d head reconstruction},
  author={Wang, Xueying and Guo, Yudong and Yang, Zhongqi and Zhang, Juyong},
  journal={IEEE Transactions on Multimedia},
  volume={24},
  pages={4028--4040},
  year={2021},
  publisher={IEEE}
}

@inproceedings{shao2024splattingavatar,
  title={Splattingavatar: Realistic real-time human avatars with mesh-embedded gaussian splatting},
  author={Shao, Zhijing and Wang, Zhaolong and Li, Zhuang and Wang, Duotun and Lin, Xiangru and Zhang, Yu and Fan, Mingming and Wang, Zeyu},
  booktitle={Proceedings of the IEEE/CVF Conference on Computer Vision and Pattern Recognition},
  pages={1606--1616},
  year={2024}
}

@inproceedings{zhang2025fate,
  title={Fate: Full-head gaussian avatar with textural editing from monocular video},
  author={Zhang, Jiawei and Wu, Zijian and Liang, Zhiyang and Gong, Yicheng and Hu, Dongfang and Yao, Yao and Cao, Xun and Zhu, Hao},
  booktitle={Proceedings of the Computer Vision and Pattern Recognition Conference},
  pages={5535--5545},
  year={2025}
}

@article{yu2021bisenet,
  title={Bisenet v2: Bilateral network with guided aggregation for real-time semantic segmentation},
  author={Yu, Changqian and Gao, Changxin and Wang, Jingbo and Yu, Gang and Shen, Chunhua and Sang, Nong},
  journal={International journal of computer vision},
  volume={129},
  pages={3051--3068},
  year={2021},
  publisher={Springer}
}

@article{wang2004image,
  title={Image quality assessment: from error visibility to structural similarity},
  author={Wang, Zhou and Bovik, Alan C and Sheikh, Hamid R and Simoncelli, Eero P},
  journal={IEEE transactions on image processing},
  volume={13},
  number={4},
  pages={600--612},
  year={2004},
  publisher={IEEE}
}

@inproceedings{athar2022rignerf,
  title={Rignerf: Fully controllable neural 3d portraits},
  author={Athar, ShahRukh and Xu, Zexiang and Sunkavalli, Kalyan and Shechtman, Eli and Shu, Zhixin},
  booktitle={Proceedings of the IEEE/CVF conference on Computer Vision and Pattern Recognition},
  pages={20364--20373},
  year={2022}
}

@inproceedings{ma20243d,
  title={3d gaussian blendshapes for head avatar animation},
  author={Ma, Shengjie and Weng, Yanlin and Shao, Tianjia and Zhou, Kun},
  booktitle={ACM SIGGRAPH 2024 Conference Papers},
  pages={1--10},
  year={2024}
}

@inproceedings{xu2023avatarmav,
  title={Avatarmav: Fast 3d head avatar reconstruction using motion-aware neural voxels},
  author={Xu, Yuelang and Wang, Lizhen and Zhao, Xiaochen and Zhang, Hongwen and Liu, Yebin},
  booktitle={ACM SIGGRAPH 2023 Conference Proceedings},
  pages={1--10},
  year={2023}
}

@article{chu2024gpavatar,
  title={GPAvatar: Generalizable and precise head avatar from image (s)},
  author={Chu, Xuangeng and Li, Yu and Zeng, Ailing and Yang, Tianyu and Lin, Lijian and Liu, Yunfei and Harada, Tatsuya},
  journal={arXiv preprint arXiv:2401.10215},
  year={2024}
}

@inproceedings{xu2024gaussian,
  title={Gaussian head avatar: Ultra high-fidelity head avatar via dynamic gaussians},
  author={Xu, Yuelang and Chen, Benwang and Li, Zhe and Zhang, Hongwen and Wang, Lizhen and Zheng, Zerong and Liu, Yebin},
  booktitle={Proceedings of the IEEE/CVF conference on computer vision and pattern recognition},
  pages={1931--1941},
  year={2024}
}

@inproceedings{zheng2022sdf,
  title={SDF-StyleGAN: implicit SDF-based StyleGAN for 3D shape generation},
  author={Zheng, Xinyang and Liu, Yang and Wang, Pengshuai and Tong, Xin},
  booktitle={Computer Graphics Forum},
  volume={41},
  pages={52--63},
  year={2022},
  organization={Wiley Online Library}
}

@inproceedings{zhuang2022mofanerf,
  title={Mofanerf: Morphable facial neural radiance field},
  author={Zhuang, Yiyu and Zhu, Hao and Sun, Xusen and Cao, Xun},
  booktitle={European conference on computer vision},
  pages={268--285},
  year={2022},
  organization={Springer}
}

@article{yao2022dfa,
  title={Dfa-nerf: Personalized talking head generation via disentangled face attributes neural rendering},
  author={Yao, Shunyu and Zhong, RuiZhe and Yan, Yichao and Zhai, Guangtao and Yang, Xiaokang},
  journal={arXiv preprint arXiv:2201.00791},
  year={2022}
}

@inproceedings{ma2023otavatar,
  title={Otavatar: One-shot talking face avatar with controllable tri-plane rendering},
  author={Ma, Zhiyuan and Zhu, Xiangyu and Qi, Guo-Jun and Lei, Zhen and Zhang, Lei},
  booktitle={Proceedings of the IEEE/CVF Conference on Computer Vision and Pattern Recognition},
  pages={16901--16910},
  year={2023}
}

@inproceedings{zielonka2025gaussian,
  title={Gaussian eigen models for human heads},
  author={Zielonka, Wojciech and Bolkart, Timo and Beeler, Thabo and Thies, Justus},
  booktitle={Proceedings of the Computer Vision and Pattern Recognition Conference},
  pages={15930--15940},
  year={2025}
}

@article{zheng2024headgap,
  title={Headgap: Few-shot 3d head avatar via generalizable gaussian priors},
  author={Zheng, Xiaozheng and Wen, Chao and Li, Zhaohu and Zhang, Weiyi and Su, Zhuo and Chang, Xu and Zhao, Yang and Lv, Zheng and Zhang, Xiaoyuan and Zhang, Yongjie and others},
  journal={arXiv preprint arXiv:2408.06019},
  year={2024}
}

@inproceedings{zielonka2025synthetic,
  title={Synthetic prior for few-shot drivable head avatar inversion},
  author={Zielonka, Wojciech and Garbin, Stephan J and Lattas, Alexandros and Kopanas, George and Gotardo, Paulo and Beeler, Thabo and Thies, Justus and Bolkart, Timo},
  booktitle={Proceedings of the Computer Vision and Pattern Recognition Conference},
  pages={10735--10746},
  year={2025}
}

@inproceedings{li2025rgbavatar,
  title={RGBAvatar: Reduced Gaussian Blendshapes for Online Modeling of Head Avatars},
  author={Li, Linzhou and Li, Yumeng and Weng, Yanlin and Zheng, Youyi and Zhou, Kun},
  booktitle={Proceedings of the Computer Vision and Pattern Recognition Conference},
  pages={10747--10757},
  year={2025}
}

@article{guo2025sega,
  title={SEGA: Drivable 3D Gaussian Head Avatar from a Single Image},
  author={Guo, Chen and Su, Zhuo and Wang, Jian and Li, Shuang and Chang, Xu and Li, Zhaohu and Zhao, Yang and Wang, Guidong and Huang, Ruqi},
  journal={arXiv preprint arXiv:2504.14373},
  year={2025}
}

\end{document}